# Suffering from Vaccines or from Government?
## - Partisan Bias in COVID-19 Vaccine Adverse Events Coverage -

TaeYoung Kang[1], Hanbin Lee[2]
[1]Underscore, [1]KAIST, [2]Seoul National University
[1]minvv23@underscore.kr, [2]hanbin973@snu.ac.kr

**Abstract**

Vaccine adverse events have been presumed to be a relatively objective measure that is immune to political polarization. The real-world data, however, shows the correlation between presidential disapproval ratings and the subjective severity of adverse events. This paper investigates the partisan bias in COVID vaccine adverse events coverage with language models that can classify the topic of vaccine-related articles and the political disposition of news comments. Based on 90K news articles from 52 major newspaper companies, we found that conservative media are inclined to report adverse events more frequently than their liberal counterparts, while the coverage itself was statistically uncorrelated with the severity of real-world adverse events. The users who support the conservative opposing party were more likely to write the popular comments from 2.3K random sampled articles on news platforms. This research implies that bipartisanship can still play a significant role in forming public opinion on the COVID vaccine even after the majority of the population's vaccination.

## 1 Introduction

The partisan bias in vaccine skepticism is well known (Olive et al., 2018; Sarathchandra et al., 2018; Motta, 2021) and is still a research topic of interest in the era of the COVID-19 pandemic (Miller 2020; Hegarty et al., 2021; Muric and Ferrara, 2021). Although there is a vague perception that the vaccine adverse events coverage including suspicious deaths, VITT (thrombocytopenia), and acute leukemia are rather more objective and scientific issues than the overall vaccine safety itself, we can still doubt the basic concept of the *vanity of objectivity* on the agenda-setting of media. For instance, sudden death after the vaccination can be described as a suspected case of vaccine side effects before causality verification is completed.

South Korea is one of the countries with the highest COVID vaccine uptake. By January 15th, 2021, 84.7% of the total population is fully vaccinated which corresponds to 94.9% of the total adult population over age 18. Despite the high vaccine uptake, its side effects and related adverse events are still great public interest and a central point of political discourse. This raises an interesting inquiry on whether political bipartisanship still plays a significant role in vaccine favorability even after the majority of the population is vaccinated.

Based on the finding that the political divide has a persistent effect on reporting of vaccine side effects even after vaccination reaches most of the population, we investigate the partisan bias in COVID vaccine adverse events coverage. Two language models that can classify the topic of the vaccine-related news and discriminate the partisan bias of users' news comment history are used. By combining these classifiers and econometric analysis, the paper analyzes the political aspects of adverse events news reporting.

## 2 Real-World Data Research Motivation

### 2.1 KDCA Region Level Data Analysis

The Korea Disease Control Agency (KCDA) collects self-reported COVID vaccine adverse events through internet and mobile messengers that are viable to everyone who received a vaccine. We retrieved regional self-reported side effect incidences of 17 provinces in South Korea through the national information disclosure system. Then we compared this regional adverse events ratio with each region's proportion of conservative opposing party's seats based on the 2020 general election.

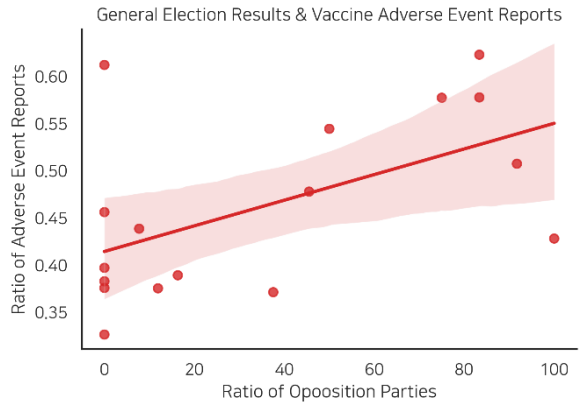

Figure 1 : Regional Level Analysis

Considering that the national vaccination policy was adopted without regional exception, there should not be a significant difference among the true value of the adverse cases among the regions. The unit increase in the seat ratio of the opposition party, however, led to a 0.0014%pt higher frequency of adverse cases report. (p<.05) The pattern was still robust when we adopted Gallup Korea's regional disapproval ratings as its substitute (p<.05)

## 2.2 Survey Data Analysis

Although the regional descriptive statistics revealed a negative correlation between the affinities with the incumbent and self-reported vaccine side effect per capita, such region-level analysis is susceptible to unobserved residual confounding and the ecological fallacy. Therefore, we proceeded to analyze individual-level data collected through a nationwide survey by Gallup Korea. We surveyed 1.5K citizens of age 18 to 69 from twelve regions across the country through stratified sampling. The overall standard error was targeted at 2.5%.

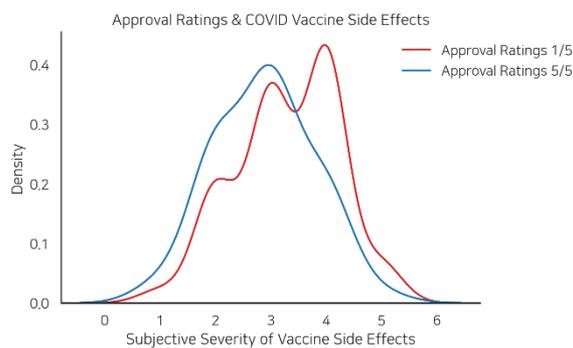

Figure 2 : Individual Level Survey Analysis

The analysis in Table 1 reconfirmed the positive correlation between the disapproval of the government (5-point Likert scale) and self-reported COVID vaccine side effect severity, showing a robust association between the two variables after adjusting age, sex, residential area, educational attainment in years and monthly income.

| subjective severity | coefficient | z-value |
|---|---|---|
| intercept | 4.214*** | 11.267 |
| residence (city=1) | -0.058 | -0.786 |
| sex (male=1) | -0.012 | -0.179 |
| marriage (married=1) | 0.152* | 1.723 |
| age | 0.007** | 2.077 |
| **approval ratings** | **-0.118*** | **-4.574** |
| income | -0.000 | -0.644 |
| educational attainment | -0.031* | -1.658 |
| No. of dosed vaccines | -0.219** | -3.387 |

***p<0.01, **p<0.05, *p<0.1
Table 1 : Robust Regression with HC3 S.E.

## 2.3 Implications

The two previous analyses on COVID vaccines revealed (1) regional-level negative correlation between the incumbency support and vaccine adverse events and (2) individual-level negative correlation between regional approval ratings and subjective severity of side effects after controlling for individual-level covariates. We do not proceed further on identifying the true causal effect of bipartisanship on self-reported vaccine side effects but proceed onto how the bipartisan divide is also observed in media coverage.

## 3 Research Objectives

Our real-world data analysis implies that political disposition significantly illustrates the report of COVID vaccine adverse events even after vaccination reaches most of the population. This phenomenon can be illustrated as a case of *motivated reasoning*, a justification of emotional prejudice that distorts the precise evaluation of the given evidence (Kunda, 1990), which is a common cognitive bias in political decision making (Nyhan & Reifler, 2010; Lodge & Taber, 2013; Leeper and Slothuus, 2014; Guay & Johnston, 2020).

However, this decision-making would not be independent of the structure of concurrent political discourse in media. Thus, based on the news data we aim to find the political bias in vaccine adverse events coverage, focusing on the following two research questions.

First, are politically conservative media more prone to write an article about COVID vaccine's adverse events?

Second, did politically conservative web portal users write more comments in news articles that deal with COVID vaccines adverse events?

To answer these questions, we would need news article data, user comments data, and language models to discriminate the topic of COVID news articles and users' political dispositions.

## 4 Data and Methods

### 4.1 Data

Korea Press Foundation (KPF) is archiving every article of major newspaper companies in South Korea since 1990. Based on its news database, we collected 90K articles written between February and December 2021, which includes the word *vaccine* in its title.

To extract the exact articles about vaccine adverse events, 6K article titles were randomly sampled and binary labeled by undergraduate and graduate three independent annotators. (0=general articles, 1=side effect and adverse events articles)

In South Korea, almost all the newspaper companies upload their articles to the web portals, since the news consumption mainly occurs in these portal websites, instead of directly visiting each media page. Naver and Daum are the most popular platforms (So et al., 2021; Han et al., 2017; Lee et al., 2015; Ryu, 2013) and people actively share their idea by writing comments below each article. To track these reactions of collected 90K articles, 39K articles that have three or more comments were matched based on the title and the name of the news provider. After the matching, 2.3K articles were random sampled from each portal, respectively, and the commenting history of the top three users per each article who earned the most recommendations to measure their partisanship.

For user partisanship classification, 25K sampled comments from Han et al (2019), 3K comments from Kang & Sim (2021), and additional manually labeled 13K comments were combined. (0=liberal / 1=neutral / 2=conservative) Since we lacked enough neutral-labeled data, 0.9K sentences from Wikipedia contents on Korean politicians were added.

Lastly, the presidential approval ratings data from Gallup Korea was used to further investigate the real-world impact of COVID vaccine adverse events coverage.

### 4.2 Methods

The Korean pretrained model KcBERT (Lee, 2020) was used for both news title and partisanship classification with a 3:1 train-test ratio with batch size 64, learning rate 1e-5, weight decay 1e-3, and 20 epochs. In the case of the 6K vaccine news title annotation set, it obtained high label accordance with 0.9434 Krippendorff's alpha, and the binary classification model achieved F1 score 0.9820. When applied to the 90K news article dataset, 7% of vaccine-related articles were exactly dealing with the issue of adverse events and side effects.

The three-class classification model for user partisanship discrimination achieved an F1 score 0.8379. After scraping each user's commenting history, 50 comments written in the political news section were randomly sampled and the maximum frequency of their predicted partisanship was assigned as a user's political orientation.

For time series analysis, econometric models including VAR (Vector Autoregression), Granger causality analysis, and impulse response analysis were adopted.

## 5 Analysis

### 5.1 Partisan Bias of News Providers

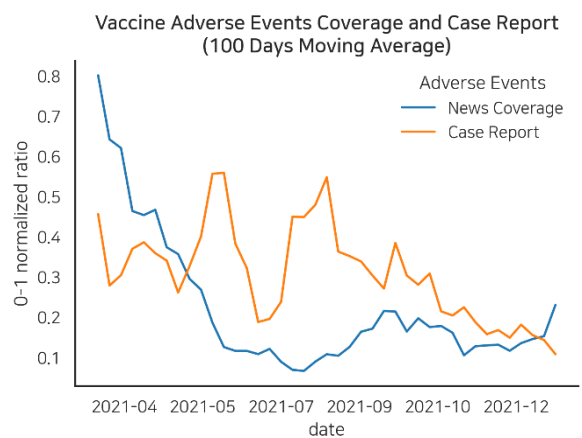

Figure 3 : Time Series Correlation Analysis

With weekly aggregated COVID19 vaccine adverse events cases, we analyzed its possible time series correlation with the actual news coverage frequency. If the correlation holds, it can be said that news media are accurately reflecting the real-world severity of adverse cases. If not, however, we could possibly doubt the motivation of vaccine agenda selection of news providers.

Based on the VAR model with the optimal lag of 4 weeks, we were not able to find any significant pattern between the number of adverse events coverage and the actual report cases. The subsequent Granger causality analysis also presented a statistical null finding, which implies that we cannot be assured of any kind of causal directions (SSR based χ2 test p-value 0.5439, likelihood ratio test p-value 0.6597)

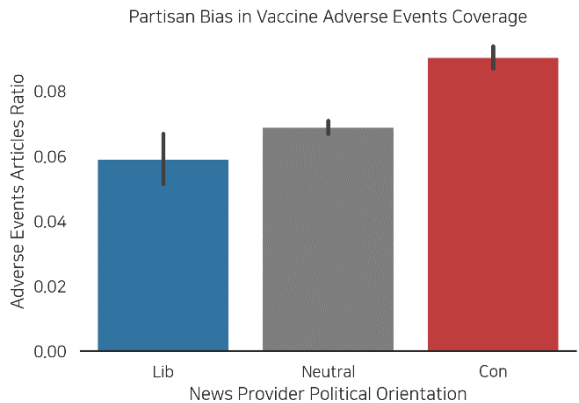

Figure 4 : Partisan Bias of News Providers

Although the adverse events news coverage was independent of the frequency of real-world events, we found the political bias among the news providers. Compared to politically moderate news providers, conservative media wrote 29.5% more articles on adverse events (p<.001) while liberal media wrote 16.3% less articles on the equivalent topic (p<.05).

### 5.2 Partisan Bias of Web Portal Users

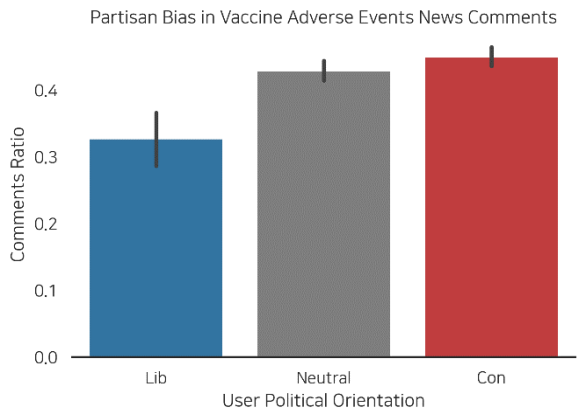

Figure 5 : Partisan Bias of Users

The political bias was also observed in user-level web portals data. Users who are predicted to be conservative opposing party supporters were more likely to write comments in adverse events news articles than Liberal incumbent supporters (p<.001). This pattern was robust even when the partisanship of news providers, the time of writing the article, and the type of portals were controlled.

| adverse events news | coefficient | t-value |
| --- | --- | --- |
| intercept | 0.538*** | 5.057 |
| **user type (cons.)** | **0.005** | **0.109** |
| **user type (lib.)** | **-0.651*** | **-6.190** |
| media type (cons.) | 0.423*** | 9.122 |
| media type (lib.) | -0.056 | -0.525 |
| portal (*Naver*=1) | -0.365*** | -4.070 |
| month | -0.111*** | -11.694 |

***p<0.01, **p<0.05, *p<0.1
Table 2 : Logistic Regression with HC3 S.E.

## 6 Conclusion

There has been a vague perception that the COVID-19 vaccine adverse events are not a political issue compared to the anti-vaxxer fake news and the partisan controversy on the general efficacy of vaccines. The real-world data, however, shows the correlation between presidential disapproval ratings and subjective severity of adverse events, implying the motivated reasoning of the critics of the incumbent government.

This paper investigates the partisan bias in COVID vaccine adverse events coverage with BERT-based language models that can classify the topic of vaccine-related articles with an F1 score 0.9820 and the political disposition of news comments with an F1 score 0.8379.

Based on 90K news articles from 52 major newspaper companies, we found that conservative media are inclined to report adverse events more frequently than their liberal counterparts, while they were independent of actual adverse reports frequencies. The users who support the conservative opposing party were also more likely to write popular comments.

This research implies that bipartisanship can still play a significant role in forming public opinion on the COVID vaccine even after the majority of the population's vaccination. Although the causal direction between the real-world phenomenon and the partisan online news media should be verified, it is likely that politically biased coverage on public health issues can distort the people's proper understanding of scientific information.

**Ethical Considerations**

All the dataset in this paper and the instruction for annotation is publicly available for replication. The pretrained language model is also accessible online, and we did not violate its intended use. The paper deanonymizes both the survey respondents and web portals users, not claiming any copyright on their comments. Also, the data annotators were informed of the usage of the dataset.